\documentclass[12pt, a4paper]{article}

\usepackage{amssymb}
\usepackage{authblk}
\usepackage[utf8]{inputenc}
\usepackage{hyperref}
\usepackage{subfig}
\usepackage{tikz}

\title{Jansen-MIDAS: a multi-level photomicrograph segmentation software
based on isotropic undecimated wavelets}

\author[a]{Alexandre Fioravante de Siqueira\thanks{Corresponding author. Phone: +55(19)3521--5362.\\
                                           \href{alexandredesiqueira@programandociencia.com}{alexandredesiqueira@programandociencia.com}}}
\author[b]{Flávio Camargo Cabrera\thanks{\href{flavioccabrera@yahoo.com.br}{flavioccabrera@yahoo.com.br}}}
\author[a]{Wagner Massayuki Nakasuga\thanks{\href{wamassa@gmail.com}{wamassa@gmail.com}}}
\author[c]{Aylton Pagamisse\thanks{\href{aylton@fct.unesp.br}{aylton@fct.unesp.br}}}
\author[b]{Aldo Eloizo Job\thanks{\href{job@fct.unesp.br}{job@fct.unesp.br}}}

\affil[a]{DRCC -- Departamento de Raios Cósmicos e Cronologia,
          IFGW -- Instituto de Física ``Gleb Wataghin'',
          UNICAMP -- University of Campinas,
          Rua Sérgio Buarque de Holanda, 777, 13083-970,
          Campinas, São Paulo, Brazil}
\affil[b]{DFQB -- Departamento de Física, Química e Biologia,
          FCT -- Faculdade de Ciências e Tecnologia,
          UNESP -- Univ Estadual Paulista,
          Rua Roberto Simonsen, 305, 19060-900,
          Presidente Prudente, São Paulo, Brazil}
\affil[c]{DMC -- Departamento de Matemática e Computação,
          FCT -- Faculdade de Ciências e Tecnologia,
          UNESP -- Univ Estadual Paulista,
          Rua Roberto Simonsen, 305, 19060-900,
          Presidente Prudente, São Paulo, Brazil}

\begin{document}

\maketitle

\clearpage

\begin{abstract}
Image segmentation, the process of separating the elements within an
image, is frequently used for obtaining information from photomicrographs.
However, segmentation methods should be used with reservations: incorrect
segmentation can mislead when interpreting regions of interest (ROI),
thus decreasing the success rate of additional procedures. Multi-Level
Starlet Segmentation (MLSS) and Multi-Level Starlet Optimal Segmentation
(MLSOS) were developed to address the photomicrograph segmentation
deficiency on general tools. These methods gave rise to Jansen-MIDAS, an
open-source software which a scientist can use to obtain a multi-level
threshold segmentation of his/hers photomicrographs. This software is presented
in two versions: a text-based version, for GNU Octave, and a graphical
user interface (GUI) version, for MathWorks MATLAB. It can be used to
process several types of images, becoming a reliable alternative to the
scientist.

\vspace{0.2cm}
\end{abstract}

\section{Introduction} 

Microscopy plays a key role in addressing several issues in
biology \cite{OLIVEIRA2014, HERCULANOHOUZEL2015}, materials
science \cite{HENZLER2015, DENG2015}, geology \cite{DEARAUJO2014}, among
other areas. Nowadays, image processing techniques are frequently used
together with microscopy, helping scientists to analyze photomicrographs
of different samples. For instance, image segmentation can be used to
separate elements within a photomicrograph, conducting the scientist to
obtain relevant information such as counting objects in a sample, or
determining their area.

There are a considerable number of computational packages available for
segmentation. One of the most known ones is the open source project Fiji
\cite{SCHINDELIN2012}, a tool based on ImageJ \cite{SCHNEIDER2012} and
aimed primarily at life sciences. Also, there are several methods rising
for segmentation of different photomicrographs, obtained by confocal
\cite{SUI2013}, magnetic resonance \cite{LIN2013}, and transmission
electron \cite{DETEMMERMAN2014} microscopies.

Despite the satisfactory segmentation which general methods present on
different images, more specific ones should be used with reservations as
they can return poor results when analyzing objects different from their
original specifications. In addition, incorrect segmentation can mislead
when interpreting regions of interest (ROI), thus decreasing the success
rate of additional computational procedures.

To address the issues on photomicrograph segmentation using general methods,
the algorithms named Multi-Level Starlet Segmentation (MLSS)
\cite{DESIQUEIRA2014a} and Multi-Level Starlet Optimal Segmentation
(MLSOS) \cite{DESIQUEIRA2014b} were developed. These methods gave rise to
the open-source software Jansen-MIDAS, available at \cite{JANSENMIDAS},
which a scientist can use to obtain a multi-level threshold segmentation
of his/hers photomicrograph, besides using supervised learning based on
the comparison between the photomicrograph ground truth and the Matthews
correlation coefficient (MCC) \cite{MATTHEWS1975} to obtain the optimal
segmentation between all levels. Jansen-MIDAS was used previously for
separating elements of two different materials, gold nanoparticles
reduced on natural rubber membranes \cite{DESIQUEIRA2014a} and fission
tracks on the surface of epidote crystals \cite{DESIQUEIRA2014b},
returning an accuracy higher than 89 \% in these applications.

This article is given as follows. Section \ref{sec:methods} presents a
brief introduction on starlet wavelet transform, Multi-Level Starlet
Segmentation (MLSS) and Multi-Level Starlet Optimal Segmentation (MLSOS)
algorithms, and Matthews correlation coefficient (MCC), the techniques
implemented in Jansen-MIDAS. Next, the Section \ref{sec:jansenmidas}
describes the two versions (text-based and graphical user interface) of
this software, and presents examples of their utilization. On Section
\ref{sec:discussion}, we discuss the use of Jansen-MIDAS on the previous
applications. Finally, in the Section \ref{sec:conclusion}, we present
our final considerations about this study.

\section{Methods}
\label{sec:methods}

Jansen-MIDAS allows the user to apply the Multi-Level Starlet Segmentation
(MLSS), a multi-level segmentation method based on starlet wavelets, and
aimed to separate elements in photomicrographs. Combining MLSS and the
Matthews correlation coefficient (MCC), we developed the Multi-Level
Starlet Optimal Segmentation (MLSOS), an optimal segmentation tool. These
methods are described in this section.

\subsection{The starlet wavelet}

The starlet wavelet is an isotropic\footnote{An isotropic wavelet is
insensitive to the orientation of features, as opposed to directional
wavelets.} and undecimated\footnote{An undecimated wavelet do not suffer
decimation (the process of reducing the sampling rate of a signal)
between its decomposition levels.} transform, suited to the analysis
of images which contains isotropic structures (e.g. astronomic
\cite{STARCK2006} or biological ones \cite{GENOVESIO2003}), and also for
structure denoising (e.g. three-dimensional electron tomographies
\cite{PRINTEMPS2016}).

The two-dimensional starlet wavelet is obtained from the scale $\phi$ and
wavelet $\psi$ functions \cite{STARCK2007, STARCK2011}:
\begin{eqnarray}
    \phi_{1D} &=& \frac{1}{12}\left(|t-2|^3-4|t-1|^3+6|t|^3-4|t+1|^3+|t+2|^3\right) \nonumber \\ %
    \phi(t_1,t_2) &=& \phi_{1D}(t_1)\phi_{1D}(t_2) \label{eq:starphi}\\ %
    \frac{1}{4}\psi\left(\frac{t_1}{2},\frac{t_2}{2}\right) &=& 
    \phi(t_1,t_2)-\frac{1}{4}\phi\left(\frac{t_1}{2},\frac{t_2}{2}\right) \label{eq:starpsi}%
\end{eqnarray}
\noindent where $\phi_{1D}$ is the one-dimensional third order B-\textit{spline}
($B_{3}$-\textit{spline}), a smooth function capable of separating large
structures within an image \cite{STARCK2010}. The wavelet function $\psi$,
in its turn, is obtained from the difference between two decomposition
levels.

Similarly to Equations \ref{eq:starphi} and \ref{eq:starpsi}, the finite
impulse response (FIR) filters $(h, g)$ related to the starlet wavelet
are defined by \cite{STARCK2010}:
\begin{eqnarray}
    h_{1D} &=& \frac{
    [
    \begin{array}{ccccc}
        1 & 4 & 6 & 4 & 1 \\ 
    \end{array}
    ]}{16} \nonumber \\ %
    h[k,l] &=& h_{1D}[k]h_{1D}[l] \label{eq:starfilth} \\ %
    g[k,l] &=& \delta[k,l]-h[k,l] \label{eq:starfiltg}
\end{eqnarray}

\noindent where $k, l = -2,...,2$, and $\delta$ is defined as $\delta[0,0] = 1$
and $\delta[k,l] = 0$ for all $[k,l] \neq (0,0)$. One can obtain the
detail wavelet coefficients from the difference between current and
previous decomposition levels, as in Equations \ref{eq:starpsi} and
\ref{eq:starfiltg}.

Using Equation \ref{eq:starfilth}, the two-dimensional starlet application
begins with a convolution between the input image $c_{0}$ and $h$
(Equation \ref{H2D}):

\begin{equation}
h = \frac{1}{16} \left[\begin{array}{c} 1 \\ 4 \\ 6 \\ 4 \\ 1 \end{array}
\right] * \frac{1}{16}\left[\begin{array}{ccccc} 1 & 4 & 6 & 4 & 1 \end{array}\right]
= \left[\begin{array}{ccccc} \frac{1}{256} & \frac{1}{64} & \frac{3}{128} & \frac{1}{64} & \frac{1}{256} \\
\frac{1}{64} & \frac{1}{16} & \frac{3}{32} & \frac{1}{16} & \frac{1}{64} \\ 
\frac{3}{128} & \frac{3}{32} & \frac{9}{64} & \frac{3}{32} & \frac{3}{128} \\
\frac{1}{64} & \frac{1}{16} & \frac{3}{32} & \frac{1}{16} & \frac{1}{64} \\ 
\frac{1}{256} & \frac{1}{64} & \frac{3}{128} & \frac{1}{64} & \frac{1}{256} \end{array}\right]
\label{H2D}
\end{equation}

This convolution returns a set of smooth coefficients corresponding to
the first starlet decomposition level, $c_{1}$. Then, the wavelet detail
coefficients for this level, $w_{1}$, are obtained from the difference
$c_{0} - c_{1}$, as discussed earlier.

\begin{figure*}[htb]
    \centering
    \includegraphics[width=0.9\textwidth]{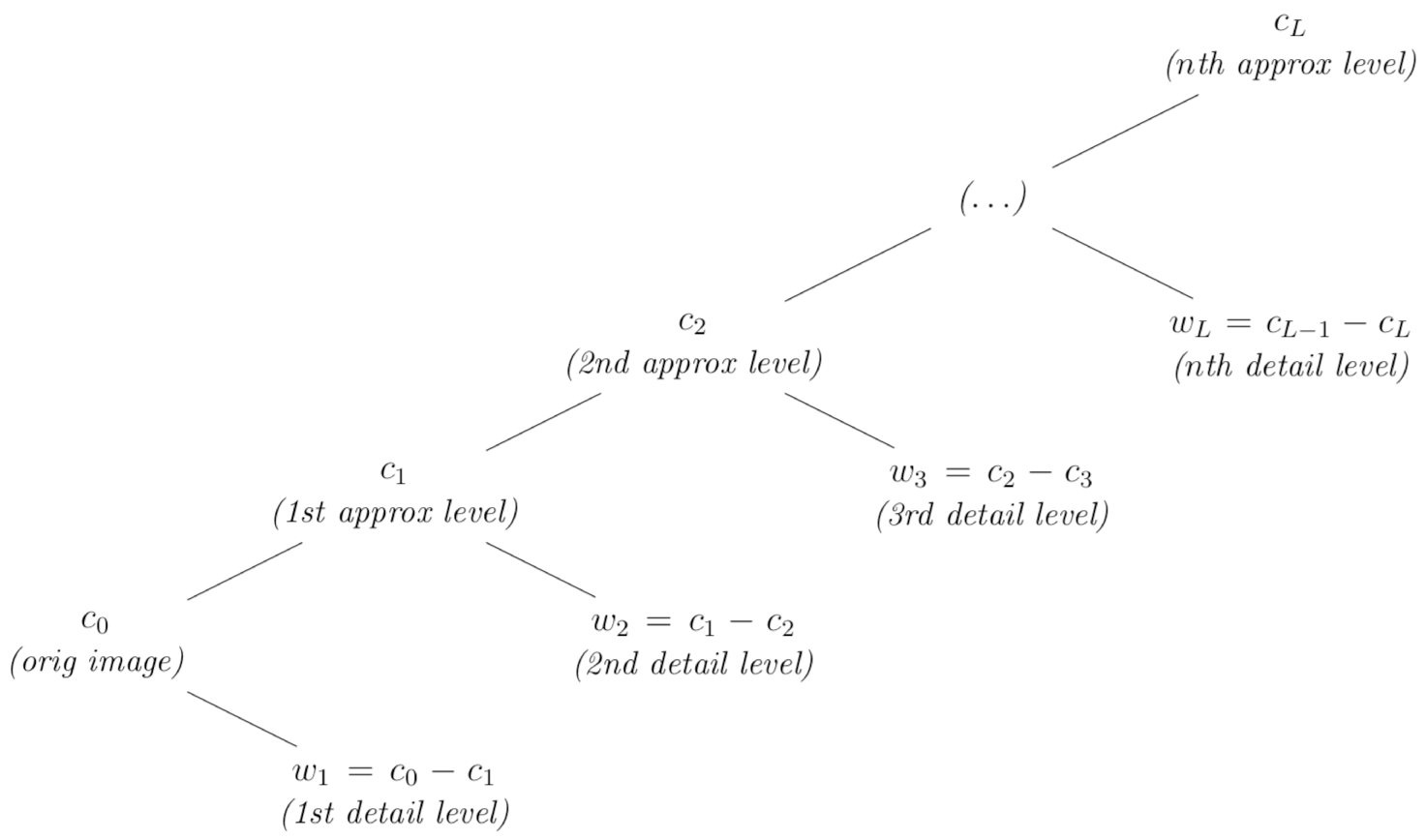}
    \caption{Tree representing the starlet decomposition structure. The
    starlet detail level $w_i$ is obtained by subtracting the
    approximation levels $c_i$ and $c_{i+1}$.}
    \label{fig:starletstruct}
\end{figure*}

Let $L$ be the last desired decomposition level. Therefore, one can
calculate the decomposition levels by:
\begin{eqnarray*}
c_{j} = c_{j-1} * h, \\
w_{j} = c_{j-1} - c_{j},
\end{eqnarray*}
\noindent where $j = 0, \ldots, L$ and $*$ is the convolution operator
(Figure \ref{fig:starletstruct}).

For $j>1$, $h$ has $2^{j}-1$ zeros between its elements \cite{MALLAT2008}.
These operations generate a set $W = \{w_{1}, \ldots, w_{L}, c_{L}\}$,
which is the \textit{starlet decomposition} of the input image $c_{0}$.

\subsection{Multi-Level Starlet Segmentation (MLSS)}

The method used in Jansen-MIDAS for photomicrograph segmentation is
denominated Multi-Level Starlet Segmentation (MLSS). It is based
on the starlet wavelet, and provides the multi-level photomicrograph
segmentation. There are two alternatives for applying MLSS: the original
and derivative algorithms.

\subsubsection{Original MLSS algorithm}

The original MLSS algorithm is based on the addition of detail levels
obtained from starlet application. After obtaining the sum of the detail
levels, the input image $c_0$ is subtracted, in order to reduce background
noise. This version is implemented as follows:
\begin{itemize}
    \item The user chooses $L$, the last desired decomposition level. The
    starlet transform is applied to the input image $c_{0}$, resulting
    in $L$ wavelet detail decomposition levels: $w_{1}, \cdots, w_{L}$.
    \item After obtaining the detail levels, the user chooses the initial
    detail level to be used in the sum,
    $L_{0}$. Detail levels lower than $L_{0}$ ($w_{1}, \cdots, w_{L_{0}-1}$)
    are ignored in this approach; this strategy can be useful for
    removing noise from the input image, which usually remains within the
    first decomposition levels.
    \item Then, the detail levels $L_{0}$ to $i$ are summed, and $c_{0}$
    is subtracted from the result:
    $$
    R_{i} = \sum_{k = L_{0}}^{i} w_{k}-c_{0},\, L_{0} \leq i \leq L,
    $$
    where $R_i$ is the starlet-related segmentation for its decomposition
    level $i$.
\end{itemize}

This algorithm results in $R_{w} = \{R_{L_{0}}, R_{L_{0}+1}, \cdots, R_{L}$\}, 
a matrix with $L-L_{0}$ segmentation levels. Each level $R_{i}$
corresponds to the starlet detail level $w_{i}$, and then the user can
choose the best segmentation level for the input image.

\subsubsection{Derivative MLSS algorithm}

The derivative MLSS algorithm also uses the starlet detail decomposition
levels, following the same basis of the original algorithm. However,
there is no subtraction of the input image, in order to preserve possible
small regions of interest (ROI). Its implementation follows:
\begin{itemize}
    \item The starlet transform is applied on the input image $c_{0}$,
    generating $L$ detail levels: $w_{1}, \cdots, w_{L}$, where $L$ is
    the last decomposition level.
    \item The user chooses the initial detail level, $L_{0}$. Detail levels
    lower than $L_{0}$ ($w_{1}, \cdots, w_{L_{0}-1}$) are ignored, and
    $w_{L_{0}}, \cdots, w_{j}$ are summed:
    $$
    R_{i} = \sum_{k = L_{0}}^{i} w_{k},\, L_{0} \leq i \leq L,
    $$
    where $R_i$ is the starlet-related segmentation for its decomposition
    level $i$.
\end{itemize}

Similarly, the algorithm returns
$R_{w} = \{R_{L_{0}}, R_{L_{0}+1}, \cdots, R_{L}$\}. The segmentation
level $R_{i}$ corresponds to the starlet detail level $w_{i}$, and the
user can choose the best segmentation level for the input image.

\subsubsection{Matthews correlation coefficient (MCC)}
\label{subsec:MCC}

Regions of interest (ROI) in an input image can be represented in a binary
image denominated ground truth (GT). Generally, the GT is obtained by an
expert, which indicates the ROI and the background on the input image.
Then, the GT can be generated representing the ROI and the background
using two different colors (usually, white and black).

One can compare the segmentation of an input image given by an
algorithm with its GT, in order to estimate the success rate of this
segmentation. From this comparison, we can define the resulting pixels
as true positives (TP), true negatives (TN), false positives (FP) and 
false negatives (FN), where:
\begin{itemize}
    \item\textbf{TP:} pixels correctly labeled as ROI by the 
    algorithm.
    \item\textbf{FP:} pixels incorrectly labeled as ROI by the 
    algorithm.
    \item\textbf{FN:} pixels incorrectly labeled as background by the 
    algorithm.
    \item\textbf{TN:} pixels correctly labeled as background by the 
    algorithm.
\end{itemize}

From TP, TN, FP and FN, one can quantify the quality of the segmentation
using the Matthews correlation coefficient (MCC) \cite{MATTHEWS1975}:
\begin{equation}
\mathcal{M} = \frac{TP*TN-FP*FN}{\sqrt{(TP+FN)(TP+FP)(TN+FP)(TN+FN)}}\times 100 \%, \label{eq:mcc}
\end{equation}

\noindent where $\mathcal{M} \in [-100 \%, 100 \%]$.

Higher $\mathcal{M}$ values indicate satisfactory segmentation: $100 \%$,
zero and $-100 \%$ represent perfect, random and opposite segmentations,
respectively \cite{BALDI2000}.

\subsubsection{Multi-Level Starlet Optimal Segmentation (MLSOS)}

In this section we present the extension of MLSS. It employs the Matthews
correlation coefficient (MCC, Equation \ref{eq:mcc}) to obtain the optimal
segmentation level. The refined method is named Multi-Level Starlet
Optimal Segmentation (MLSOS), and is defined as follows
\cite{DESIQUEIRA2014b,DESIQUEIRA2014c}:

\begin{itemize}
    \item MLSS is applied in a input image $c_0$ that has its GT for $L$
    desired starlet decomposition levels, thus acquiring
    $R_w = \{R_{L_{0}}, R_{L_{0}+1}, \cdots, R_{L}\}$.
    \item The segmentation results for each starlet level, $R_{i}$, with 
    $i = L_{0}, \cdots, L$, are compared with the GT of the input 
    image, thereby obtaining TP$_{i}$, TN$_{i}$, FP$_{i}$ and 
    FN$_{i}$.
    \item Based on these values, MCC is calculated for each $R_{i}$.
\end{itemize}

Therefore, the optimal segmentation level obtained for the input
photomicrograph is the one which returns the higher MCC value between
segmentation levels $R_{i}, L_{0} \leq i \leq L$ obtained by MLSS.

Using MLSOS, one can establish the optimal segmentation level for the
photomicrographs of a sample representing the set, thus estimating the
optimal level for the entire photomicrograph set.

\section{Jansen-MIDAS description and operating instructions}
\label{sec:jansenmidas}

The software Jansen-MIDAS\footnote{The name Jansen-MIDAS is a tribute to
the (possible) microscope inventors, Zacharias Jansen, and his father Hans,
followed by an acronym: Microscopic Data Analysis Software.} contains
implementations for the MLSS and MLSOS methods. Using this software,
scientists can apply these techniques on their own photomicrographs.

There are two versions of Jansen-MIDAS: one is based on text input,
built for GNU Octave, and another has a graphical user interface (GUI),
aimed to MATLAB users\footnote{MATLAB has a tool for creating graphical user
interfaces, named GUIDE. Using it, the programmer can easily develop a
GUI to integrate his/hers functions. Unfortunately, during the conception of
this study, there was no equivalent tool on GNU Octave, difficulting the
development of GUIs on that language.}. They are distributed in the same
package, in the folders \texttt{TEXTMODE} and \texttt{GUIMODE},
containing the text-based version and the GUI version respectively. On
the next section, we describe both versions and how to use them.

\subsection{Text-based version}

Using Jansen-MIDAS's text-based version is straightforward. After starting
GNU Octave, it presents its GUI containing the \textit{Command Window}
and its prompt, represented by two ``greater than or equal'' symbols
(\texttt{>>}). If Octave is running from the \texttt{TEXTMODE} folder,
Jansen-MIDAS is started typing the following command in the Command
Window:

\noindent\texttt{>> jansenmidas();}

Alternatively, the user can inform the variables which will store the
processing results on Octave. This can be done using the following
command, which starts Jansen-MIDAS and stores the processing results on
the variables \texttt{D}, \texttt{R}, \texttt{COMP} and \texttt{MCC}:

\noindent\texttt{>> [D,R,COMP,MCC] = jansenmidas();}

These variables represent:
\begin{itemize}
    \item\textbf{\texttt{D}:} the starlet detail decomposition levels.
    \item\textbf{\texttt{R}:} the MLSS segmentation levels.
    \item\textbf{\texttt{COMP}:} a color comparison between the input
    photomicrograph and its GT, representing TP, FP, and FN pixels.
    \item\textbf{\texttt{MCC}:} the Matthews correlation coefficient
    values for each segmentation level.
\end{itemize}

An example of applying the text-based version follows. First, the
software presents itself:\\

\noindent\texttt{\%\%\%\%\%\%\%\%\%\%\%\%\%\%\%\%\%\%\%\%\%\%\%\%\%\%\%\%\%\%\%\%\%\%\%\%\%\%\%\%\%\%\%\%\%\%\%\%\%\%\%\%\%\%\%\%\\
\%\%\%\%\%\%\%\%\%\%\%\%\%\%\% Welcome to Jansen-MIDAS \%\%\%\%\%\%\%\%\%\%\%\%\%\%\%\%\\
\%\%\%\%\%\%\%\%\%\% Microscopic Data Analysis Software \%\%\%\%\%\%\%\%\%\%\\
\%\%\%\%\%\%\%\%\%\%\%\%\%\%\%\%\%\%\%\%\%\%\%\%\%\%\%\%\%\%\%\%\%\%\%\%\%\%\%\%\%\%\%\%\%\%\%\%\%\%\%\%\%\%\%\%\\
}

The first step is to provide the first starlet detail level to be
considered on the segmentation. If the user enters $3$, for example,
the starlet detail levels $1$ and $2$ will be disregarded from the
segmentation. Remind that lower and higher detail levels present smaller
and larger detail ROI, respectively.

\noindent\texttt{Initial detail level to consider in segmentation:}

In this example we suppose the user does not input a first detail level.
When this happens, Jansen-MIDAS assumes this value as $1$:

\noindent\texttt{Assuming default value, initial level equals 1. Continue...}

After that, the software asks the last desired segmentation level.

\noindent\texttt{Last detail level to consider in segmentation:}

Similarly to when choosing the first level, if the user does not input
the last segmentation, the software assumes the default value, $5$.
In this example, we consider the last segmentation level equals to $3$:

\noindent\texttt{Last detail level to consider in segmentation: 3}

\begin{figure*}[htb]
    \centering
    \subfloat[Figure \texttt{test1.jpg}, suitable for testing the original
    MLSS.]
    {
    \label{subfig:test1}
    {\includegraphics[width=0.48\textwidth]{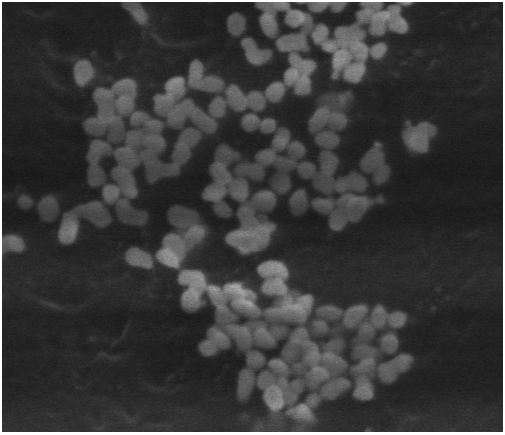}}
    }
    ~
    \subfloat[Figure \texttt{test2.jpg}, suitable for testing the derivative
    MLSS.]
    {
    \label{subfig:test2}
    {\includegraphics[width=0.48\textwidth]{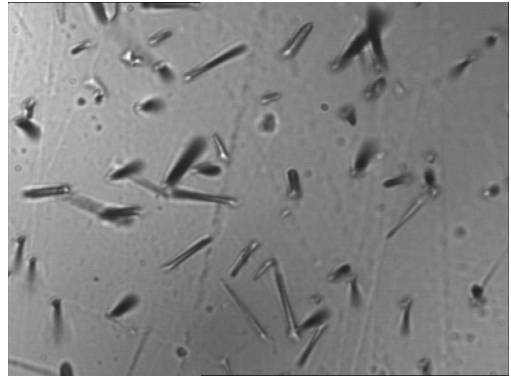}}
    }
    \caption{Test images provided with Jansen-MIDAS. These pictures are
    suitable for testing both original and derivative MLSS algorithms.}
    \label{fig:testimages}
\end{figure*}

Now Jansen-MIDAS asks the name of the photomicrograph to be processed.
The \texttt{TEXTMODE} folder contains two test images, named
\texttt{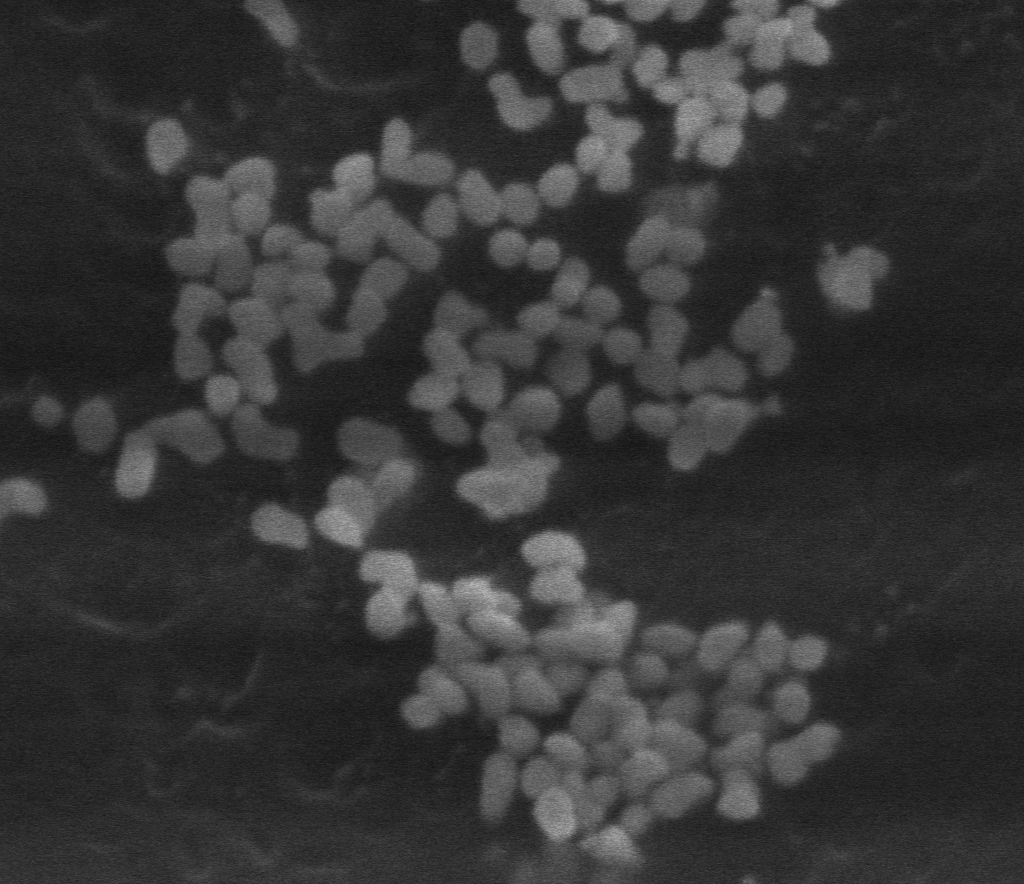} and \texttt{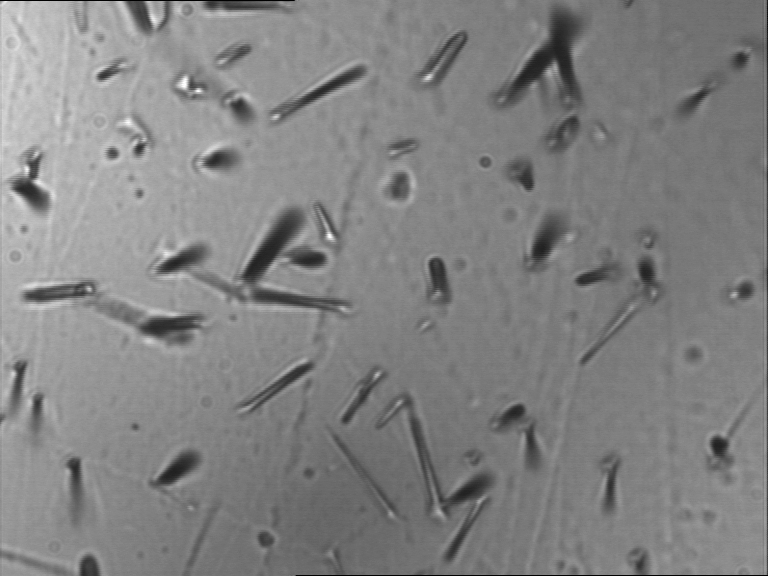} (Figure \ref{fig:testimages}).
These images are suitable to test the original and derivative MLSS
algorithms, respectively.

\noindent\texttt{Please type the original image name:}

For this example, we chose the image \texttt{test1.jpg}\footnote{The optimal
segmentation for this photomicrograph is obtained using $L_0 = 3$ and $L = 7$
\cite{DESIQUEIRA2014a}.}. Then, the software performs the MLSS on it,
asking before what processing algorithm to use (original or derivative).

\noindent\texttt{Applying MLSS...\\
Type V for Variant or any to Original MLSS:}

In order to use the derivative algorithm, the user will type \texttt{v}
or \texttt{V}. However, the original algorithm is the suitable one for
the photomicrograph \texttt{test1.jpg}. To apply it, simply press
\texttt{Enter}.

After that, the software asks if it will apply MLSOS on the
photomicrograph; if yes, it is sufficient to type \texttt{y} or \texttt{Y}.
In order to do that, the user should have the GT image representing the
ROI on the input photomicrograph, thus Jansen-MIDAS will estimate the
MCC values for each segmentation and generate the comparison between
input image and GT.

\noindent\texttt{Do you want to apply MLSOS (uses GT image)?\\
Please type a GT image name:}

\begin{figure*}[htb]
    \centering
    \subfloat[Figure \texttt{test1GT.jpg}, GT image of Figure
    \ref{subfig:test1}.]
    {\label{subfig:test1gt}
    {\includegraphics[width=0.48\textwidth]{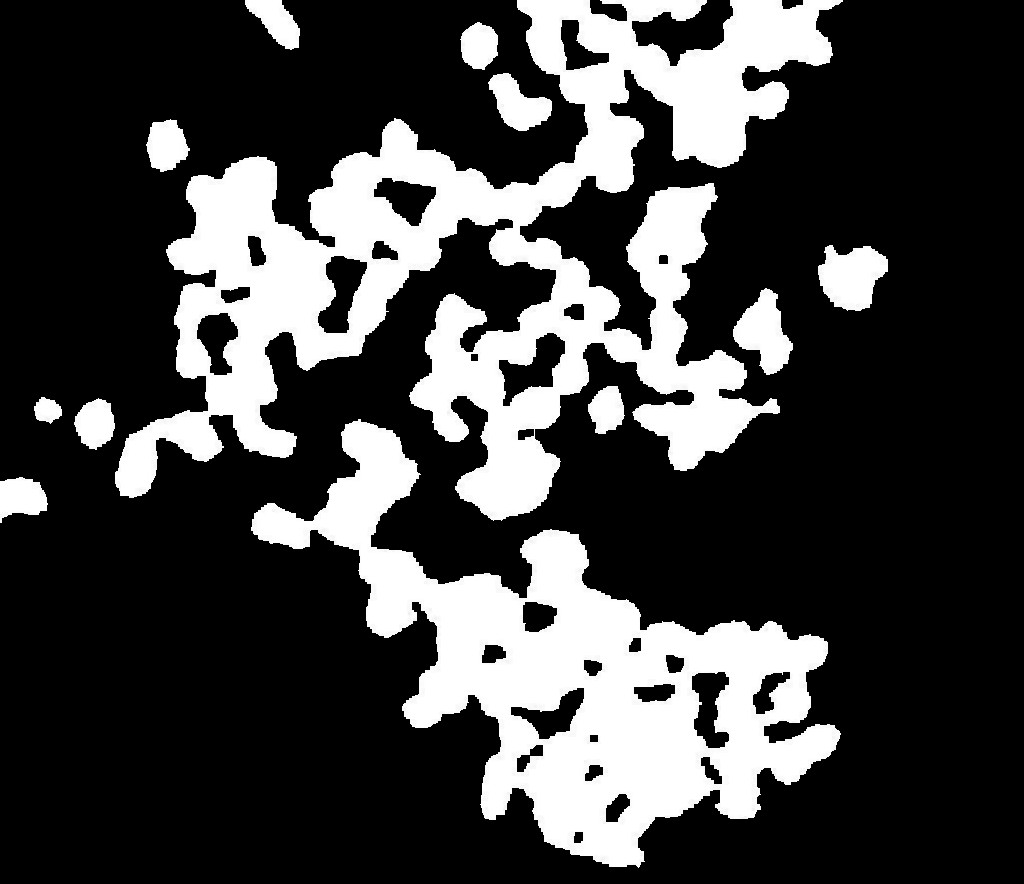}}
    }
    ~
    \subfloat[Figure \texttt{test2GT.jpg}, GT image of Figure
    \ref{subfig:test2}.]
    {\label{subfig:test2gt}
    {\includegraphics[width=0.48\textwidth]{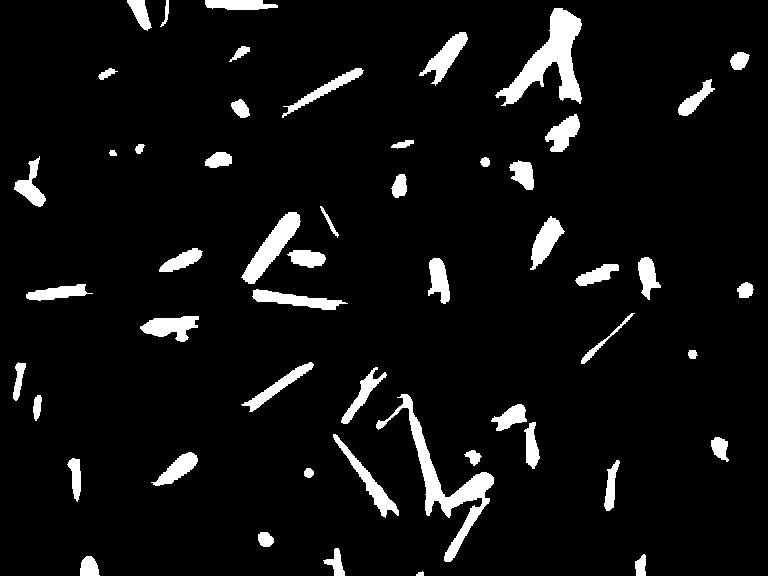}}
    }
    \caption{\textit{Ground truth} (GT) of the test photomicrographs 
    \texttt{text1.jpg} and \texttt{text2.jpg}, also provided with the
    \textit{software} Jansen-MIDAS. Together with their respective original
    images, the GT are used to apply MLSOS.}
    \label{fig:testgt}
\end{figure*}

The folder \texttt{TEXTMODE} contains also the GT images of the test
photomicrographs. The file names are \texttt{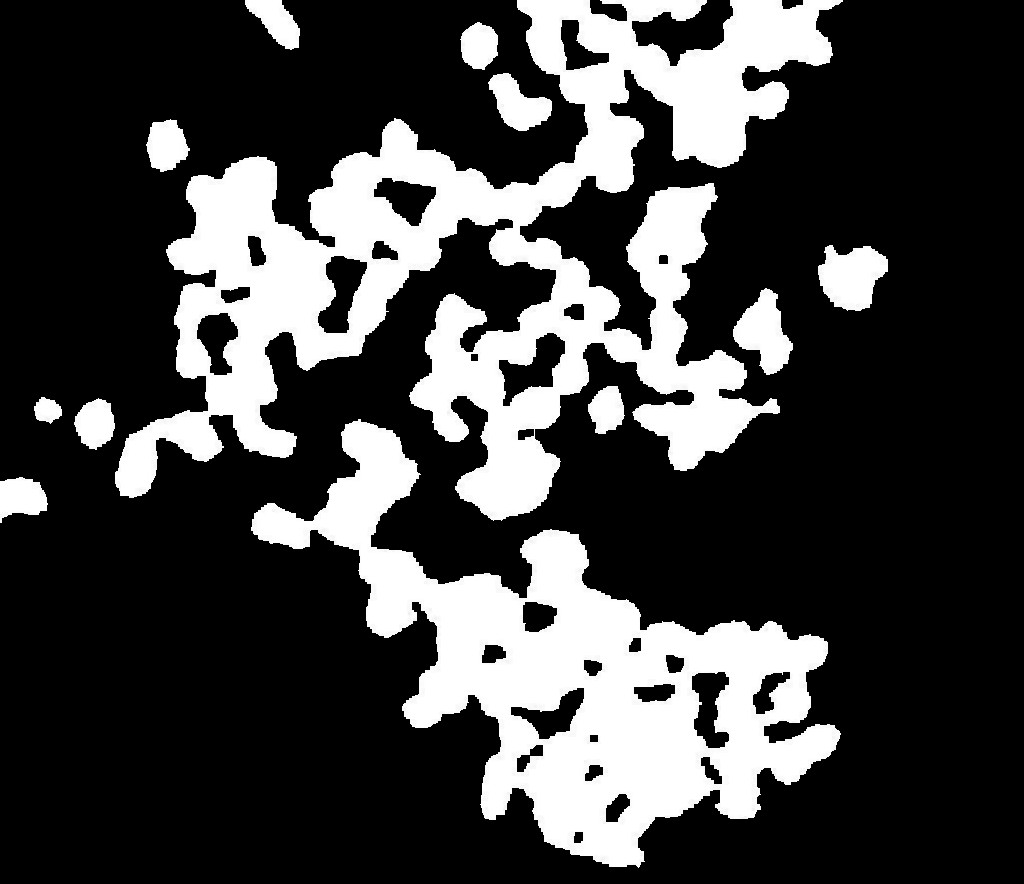} and
\texttt{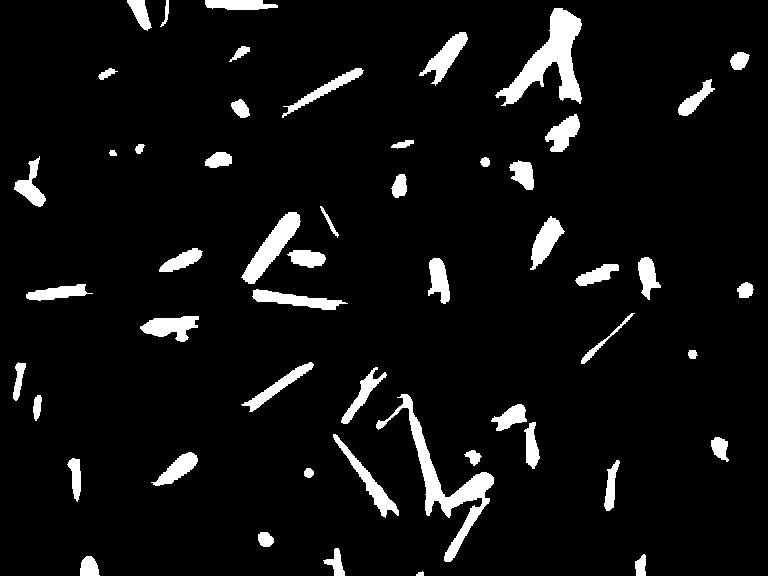} (Figure \ref{fig:testgt}). Since we chose
\texttt{test1.jpg}, \texttt{test1GT.jpg} is the GT image to be used on
the comparison.

\begin{figure*}[htb]
    \centering
    \includegraphics[width=0.8\textwidth]{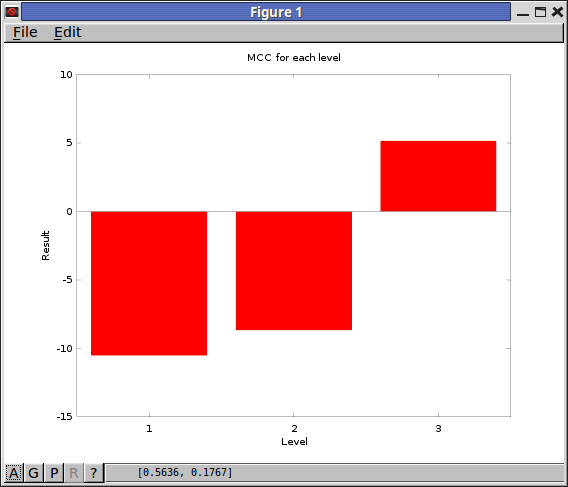}
    \caption{The Matthews correlation coefficient plot presented by
    Jansen-MIDAS's text-based version, using \texttt{test1.jpg} and 
    \texttt{test1GT.jpg}. First and last segmentation levels equal to 
    $1$ and $3$, respectively.}
    \label{fig:test1mcc}
\end{figure*}

When MLSOS application finishes, the values of MCC for each segmentation
level are shown on a plot (Figure \ref{fig:test1mcc}). Then the program
asks if it should record the resulting images or only display them.

\noindent\texttt{Type Y to save images or any to show them:}

If the user enters \texttt{y} or \texttt{Y}, the images will be stored
with the same name of the original image, and information about the
results, in three groups:

\begin{enumerate}
    \item\texttt{D}, representing the starlet detail decomposition levels;
    \item\texttt{R}, representing the MLSS segmentation levels;
    \item\texttt{COMP}, representing the comparison between the input
    photomicrograph and its GT, when the last is provided.
\end{enumerate}

For first and last levels equal to $1$ and $3$, the information
presented on the screen is:

\noindent\texttt{Saving detail image... Level: 1\\
Saving detail image... Level: 2\\
Saving detail image... Level: 3\\
Saving segmentation image... Level: 1\\
Saving segmentation image... Level: 2\\
Saving segmentation image... Level: 3\\
Saving comparison image... Level: 1\\
Saving comparison image... Level: 2\\
Saving comparison image... Level: 3
}

\begin{figure*}[htb]
    \centering
    \subfloat[Window \texttt{D1}.]
    {\label{subfig:test1d1}
    {\includegraphics[width=0.32\textwidth]{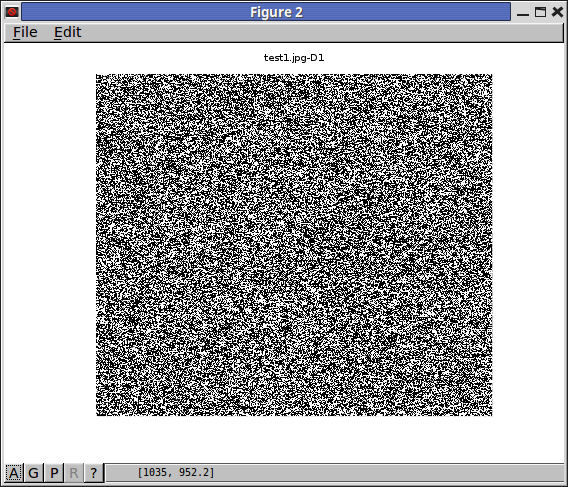}}}
    ~
    \subfloat[Window \texttt{D2}.]
    {\label{subfig:test1d2}
    {\includegraphics[width=0.32\textwidth]{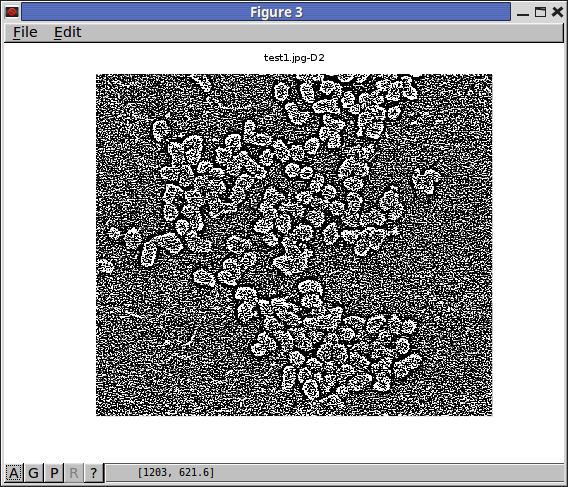}}}
    ~
    \subfloat[Window \texttt{D3}.]
    {\label{subfig:test1d3}
    {\includegraphics[width=0.32\textwidth]{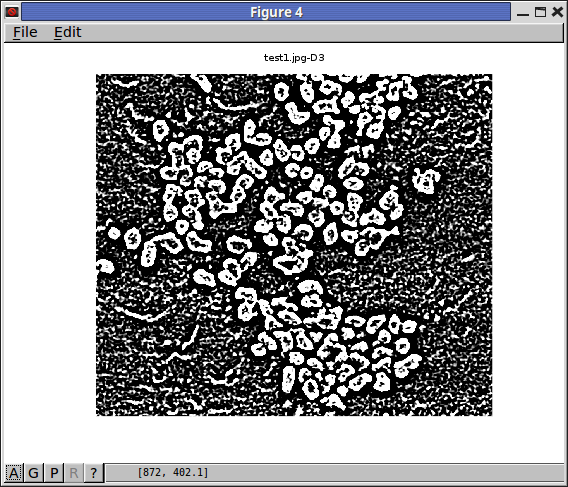}}}\\
    \subfloat[Window \texttt{R1}.]
    {\label{subfig:test1r1}
    {\includegraphics[width=0.32\textwidth]{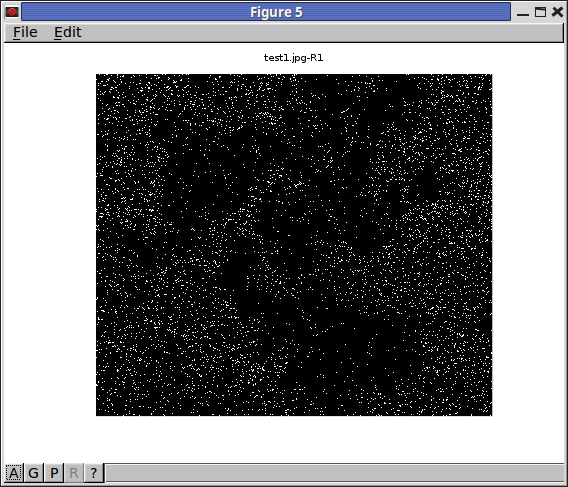}}}
    ~
    \subfloat[Window \texttt{R2}.]
    {\label{subfig:test1r2}
    {\includegraphics[width=0.32\textwidth]{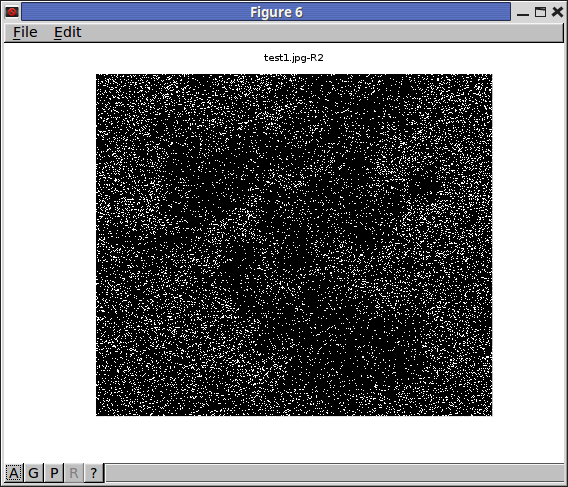}}}
    ~
    \subfloat[Window \texttt{R3}.]
    {\label{subfig:test1r3}
    {\includegraphics[width=0.32\textwidth]{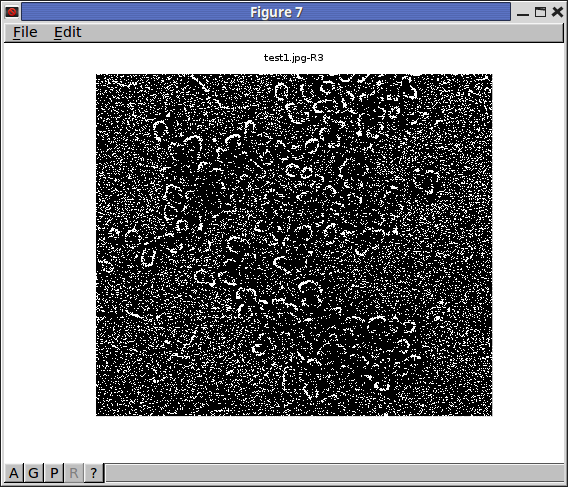}}}\\
    \subfloat[Window \texttt{COMP1}.]
    {\label{subfig:test1comp1}
    {\includegraphics[width=0.32\textwidth]{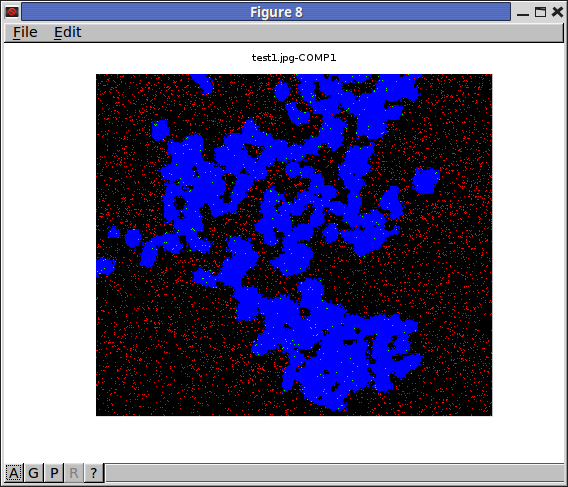}}}
    ~
    \subfloat[Window \texttt{COMP2}.]
    {\label{subfig:test1comp2}
    {\includegraphics[width=0.32\textwidth]{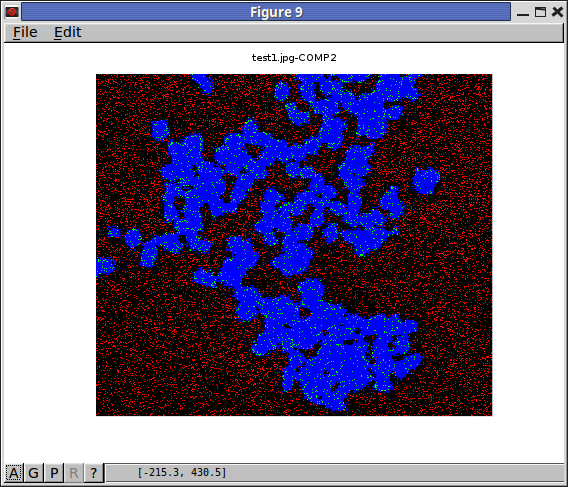}}}
    ~
    \subfloat[Window \texttt{COMP3}.]
    {\label{subfig:test1comp3}
    {\includegraphics[width=0.32\textwidth]{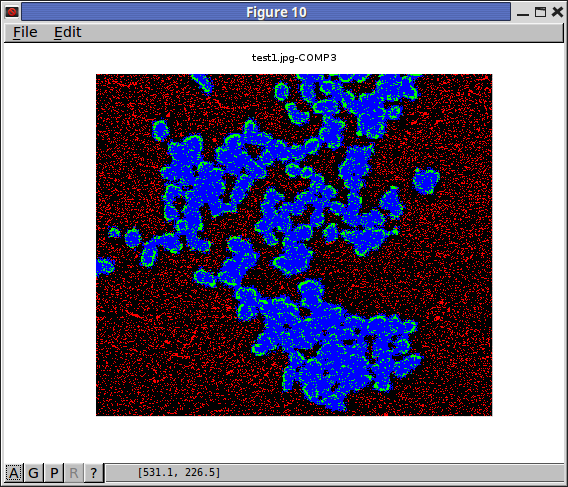}}}\\
    \caption{Windows presenting Jansen-MIDAS's application results on the
    text-based version, using \texttt{test1.jpg} and \texttt{test1GT.jpg}.
    The application of MLSOS returns three image sets: \textit{detail}
    (\texttt{D}), \textit{segmentation} (\texttt{R}) e \textit{comparison}
    (\texttt{COMP}). The segmentation levels are $1$ (first) and $3$
    (last), generating three images on each set.}
  \label{fig:test1results}
\end{figure*}

When the user chooses to not store the images, the results are presented
directly on the screen (Figure \ref{fig:test1results}). For first and
last starlet detail levels equal to $1$ and $3$, the presented information
follows:

\noindent\texttt{Showing detail image... Level: 1\\
Showing detail image... Level: 2\\
Showing detail image... Level: 3\\
Showing segmentation image... Level: 1\\
Showing segmentation image... Level: 2\\
Showing segmentation image... Level: 3\\
Showing comparison image... Level: 1\\
Showing comparison image... Level: 2\\
Showing comparison image... Level: 3
}

\clearpage

Finally, Jansen-MIDAS processing ends with the following final message:

\noindent\texttt{End of processing. Thanks!}

\subsection{Graphical user interface version}

Jansen-MIDAS's graphical version is also easy to use, although it is not
straightforward as the text-based version. This version has interactive
elements such as buttons, check and text boxes.

As in GNU Octave, the MATLAB environment has several areas. Its prompt
is also available on the \textit{Command Window} and represented by
\texttt{>>}. Inside the folder \texttt{GUIMODE}, Jansen-MIDAS starts
when typing the following command in MATLAB:

\noindent\texttt{>> JansenMIDAS}

\begin{figure*}[htb]
    \centering
    \includegraphics[width=0.8\textwidth]{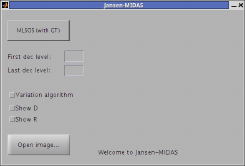}
    \caption{Jansen-MIDAS's graphical user interface (GUI), developed
    for MathWorks MATLAB.}
    \label{fig:jansenmidasgui}
\end{figure*}

Then the initial screen is presented (Figure \ref{fig:jansenmidasgui}).
The first elements presented are:
\begin{itemize}
    \item The state button \textit{MLSOS (with GT)}. When pressed, the
    software will apply MLSOS besides MLSS.
    \item The text boxes \textit{First dec level} and \textit{Last dec 
    level}, where the user can inform to Jansen-MIDAS what are the first
    and the last segmentation levels to use.
    \item The check box \textit{Variation algorithm}. When checked, the
    software applies the derivative MLSS algorithm.
    \item The check boxes \textit{Show D} and \textit{Show R}. When
    checked, Jansen-MIDAS will present the starlet detail levels and
    segmentation results instead of storing them on the disk.
    \item The welcome text \textit{Welcome to Jansen-MIDAS}, which
    presents information on the processing as the software performs its
    tasks.
    \item The button \textit{Open image...}, which asks the input image,
    to the user, thus starting the processing.
\end{itemize}

One example of Jansen-MIDAS's application using the GUI version is given
below. The processing starts when the user clicks on the button
\textit{Open image...}. Then, a window asks the input photomicrograph.
The \texttt{GUIMODE} folder also contains the test images \texttt{test1.jpg}
and \texttt{test2.jpg}, which can be used to simulate the software
usage (Figure \ref{fig:testimages}).

\begin{figure*}[htb]
    \centering
    \includegraphics[width=0.8\textwidth]{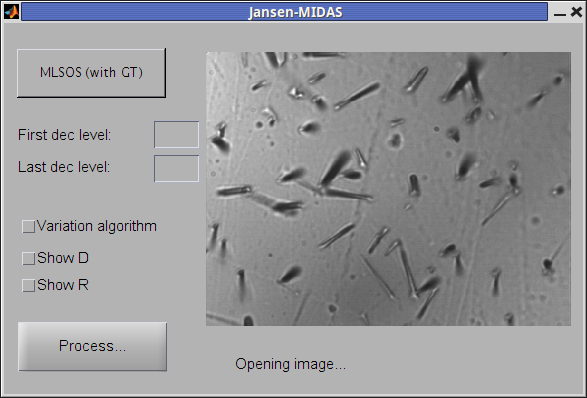}
    \caption{Jansen-MIDAS presenting the image to be segmented. After
    choosing the input photomicrograph, the button \textit{Open image...}
    becomes \textit{Process...}, and the welcome text is changed to
    \textit{Opening image...}.}
    \label{fig:jansenmidasgui2}
\end{figure*}

In this example we suppose the chosen photomicrograph is
\texttt{test2.jpg}\footnote{The optimal segmentation for this
photomicrograph is obtained using $L_0 = 3$ and $L = 7$ \cite{DESIQUEIRA2014b}.}.
The welcome text indicates the performing action, being changed to
\textit{Opening image...}. Then, the software shows the input image and
the button \textit{Open image...} becomes \textit{Process...} (Figure
\ref{fig:jansenmidasgui2}).

At this point the user can choose other options:
\begin{itemize}
    \item If MLSOS will be used, pressing the button \textit{MLSOS (with
    GT)}.
    \item The first and last segmentation levels, on the text boxes
    \textit{First dec level} and \textit{Last dec level}. As in
    Jansen-MIDAS's text version, the default first and last segmentation
    levels, assumed when these boxes do not receive values, are $1$ and
    $5$ respectively.
    \item If the software will present the starlet detail levels and
    segmentation results on the screen, using the check boxes
    \textit{Show D} and \textit{Show R}.
\end{itemize}

\begin{figure*}[htb]
    \centering
    \includegraphics[width=0.8\textwidth]{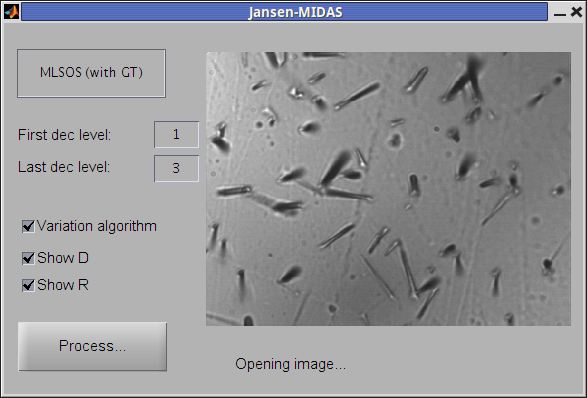}
    \caption{Jansen-MIDAS's GUI interface corresponding to applying
    MLSOS, first and last segmentation levels equal to $1$ and $3$
    respectively, using the derivative MLSS algorithm and presenting
    starlet detail levels (\texttt{D}) and segmentation results
    (\texttt{R}).}
    \label{fig:jansenmidasgui3}
\end{figure*}

The derivative MLSS algorithm is suited to segment \texttt{test2.jpg}.
Thus, if the user would want to apply MLSOS, with first and last
segmentation levels equal to $1$ and $3$, derivative MLSS algorithm and
showing starlet details (\texttt{D}) and segmentation results (\texttt{R}),
the options on the software interface would be filled as in Figure
\ref{fig:jansenmidasgui3}

After choosing the desired options, the processing starts when the button
\textit{Process...} is pressed. In this example the user pressed the state
button \textit{MLSOS (with GT)}; therefore, Jansen-MIDAS asks the GT image
corresponding to the input photomicrograph. The folder \texttt{GUIMODE}
also contains the GT images of the test photomicrographs, \texttt{test1GT.jpg}
and \texttt{test2GT.jpg} (Figure \ref{fig:testgt}). The GT image
\texttt{test2GT.jpg} is used to apply MLSOS on \texttt{test2.jpg}.

\begin{figure*}[htb]
    \centering
    \includegraphics[width=0.8\textwidth]{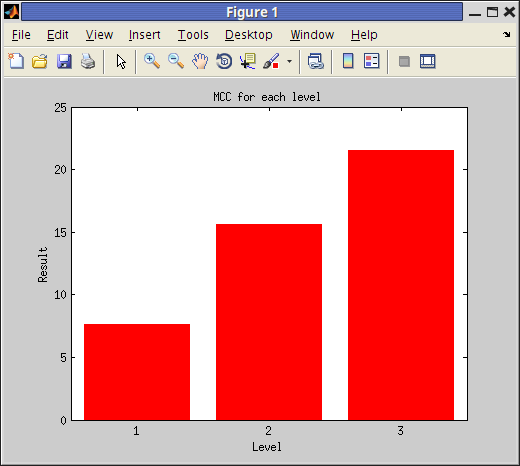}
    \caption{Matthews correlation coefficient presented by Jansen-MIDAS's
    application, GUI mode, using \texttt{test2.jpg} and \texttt{test2GT.jpg}.
    First and last segmentation levels equal to $1$ and $3$ respectively.}
    \label{fig:test2mcc}
\end{figure*}

\begin{figure*}[htb]
    \centering
    \subfloat[Window \texttt{D1}.]
    {\label{subfig:test2d1}
    {\includegraphics[width=0.32\textwidth]{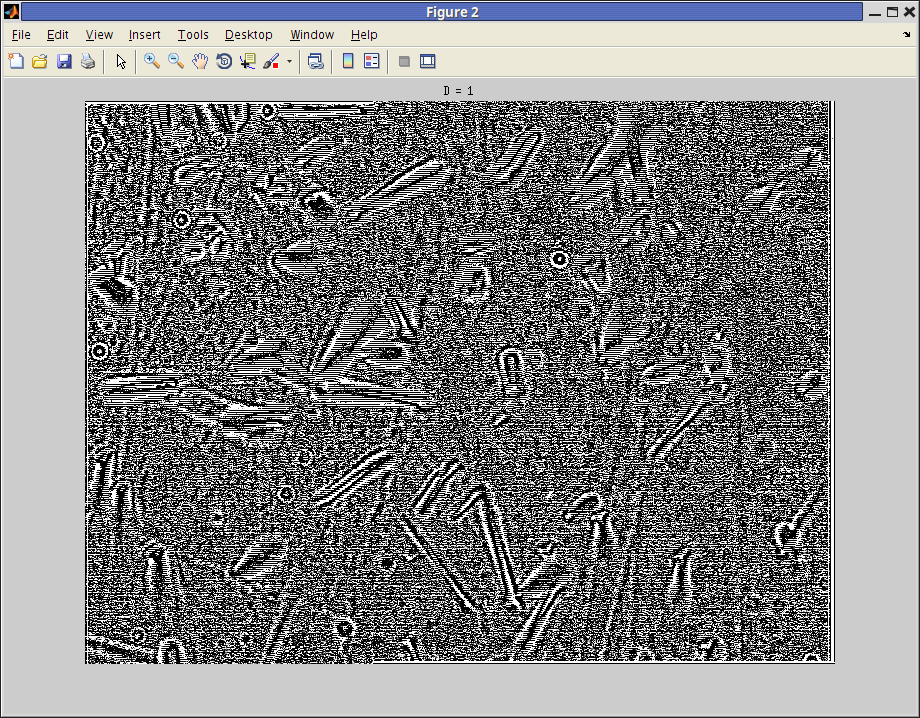}}}
    ~
    \subfloat[Window \texttt{D2}.]
    {\label{subfig:test2d2}
    {\includegraphics[width=0.32\textwidth]{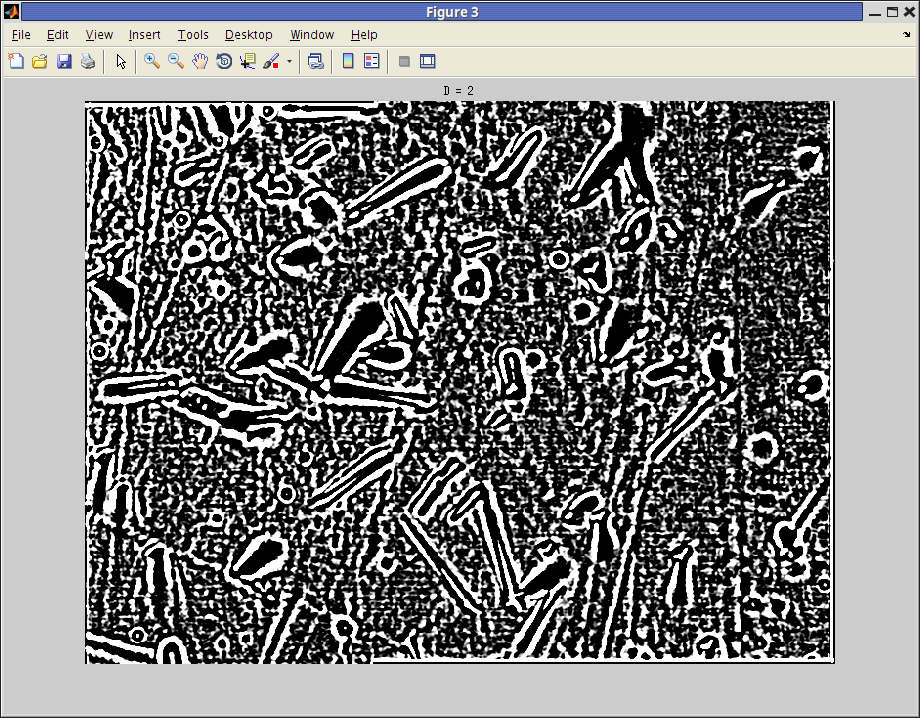}}}
    ~
    \subfloat[Window \texttt{D3}.]
    {\label{subfig:test2d3}
    {\includegraphics[width=0.32\textwidth]{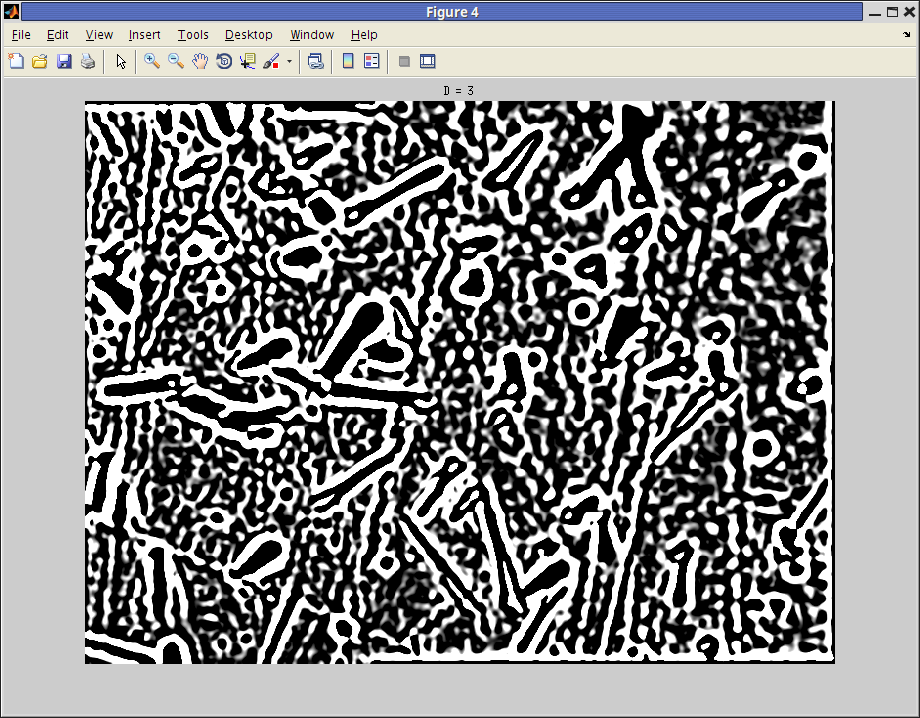}}}\\
    \subfloat[Window \texttt{R1}.]
    {\label{subfig:test2r1}
    {\includegraphics[width=0.32\textwidth]{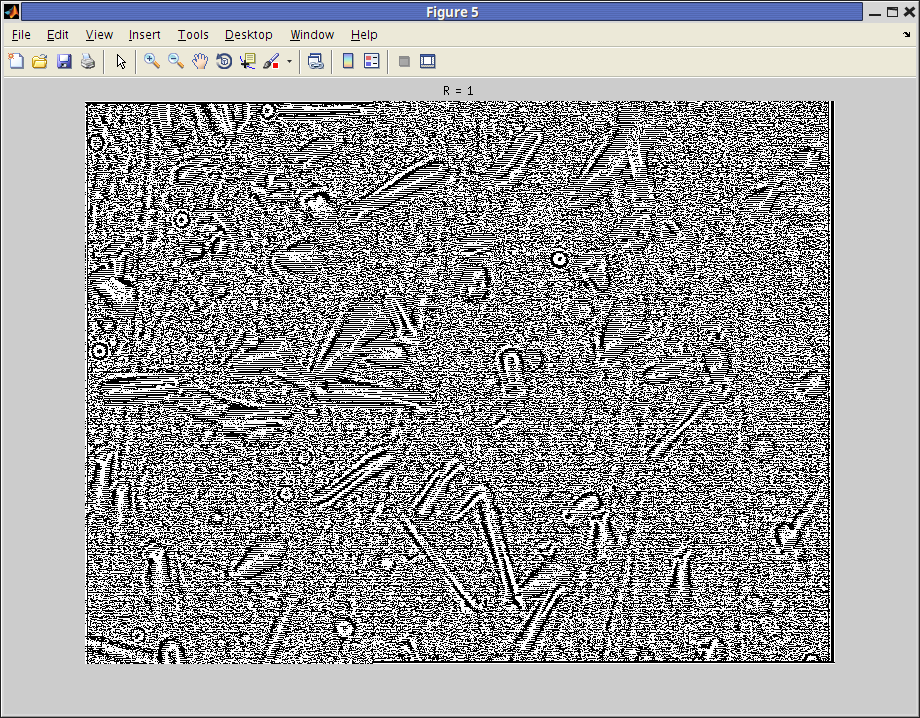}}}
    ~
    \subfloat[Window \texttt{R2}.]
    {\label{subfig:test2r2}
    {\includegraphics[width=0.32\textwidth]{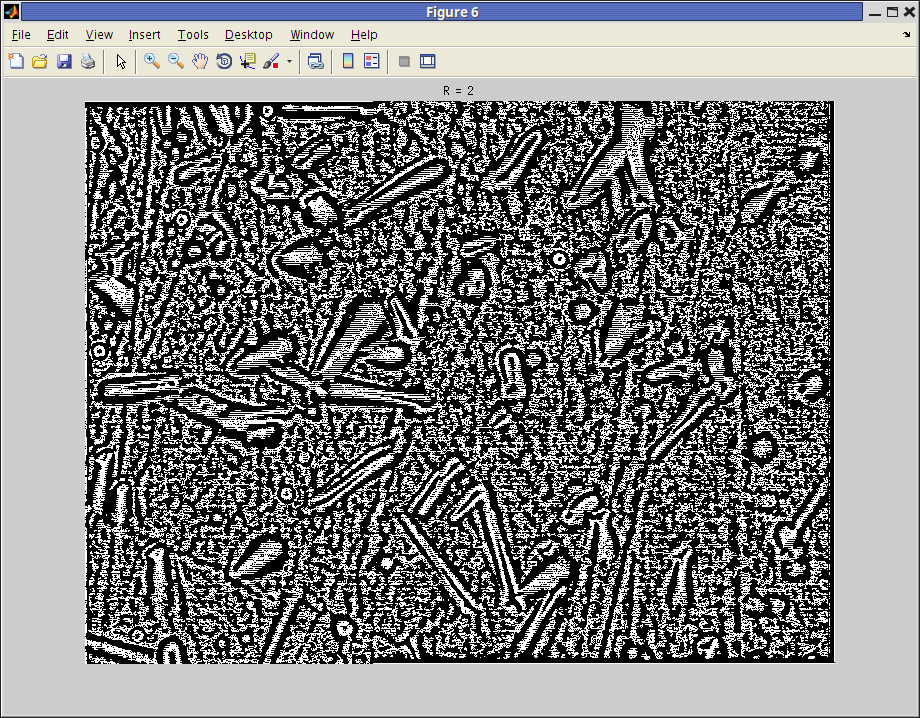}}}
    ~
    \subfloat[Window \texttt{R3}.]
    {\label{subfig:test2r3}
    {\includegraphics[width=0.32\textwidth]{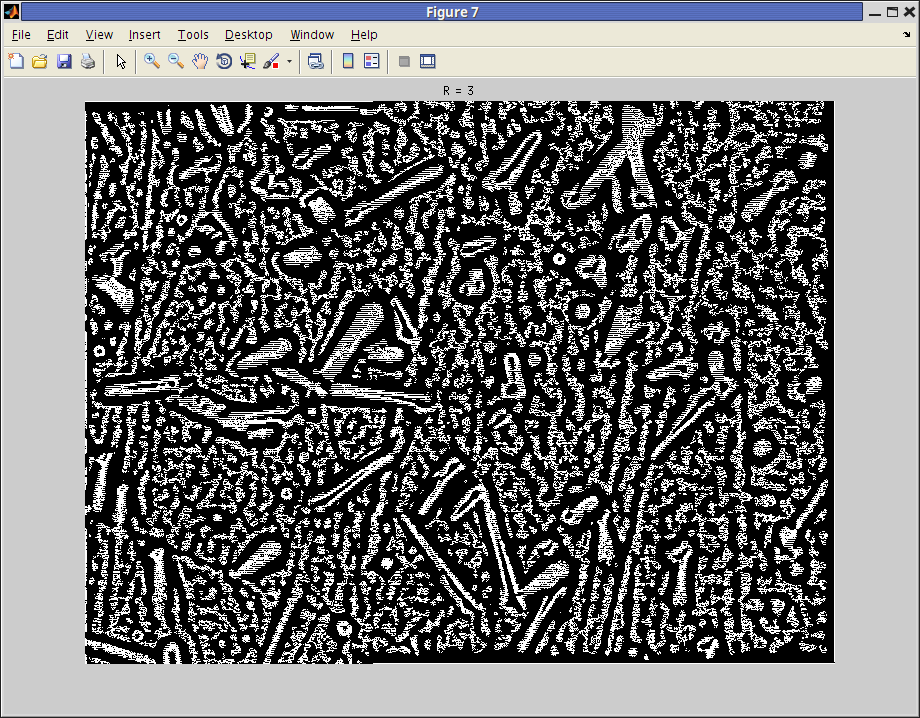}}}\\
    \subfloat[Window \texttt{COMP1}.]
    {\label{subfig:test2comp1}
    {\includegraphics[width=0.32\textwidth]{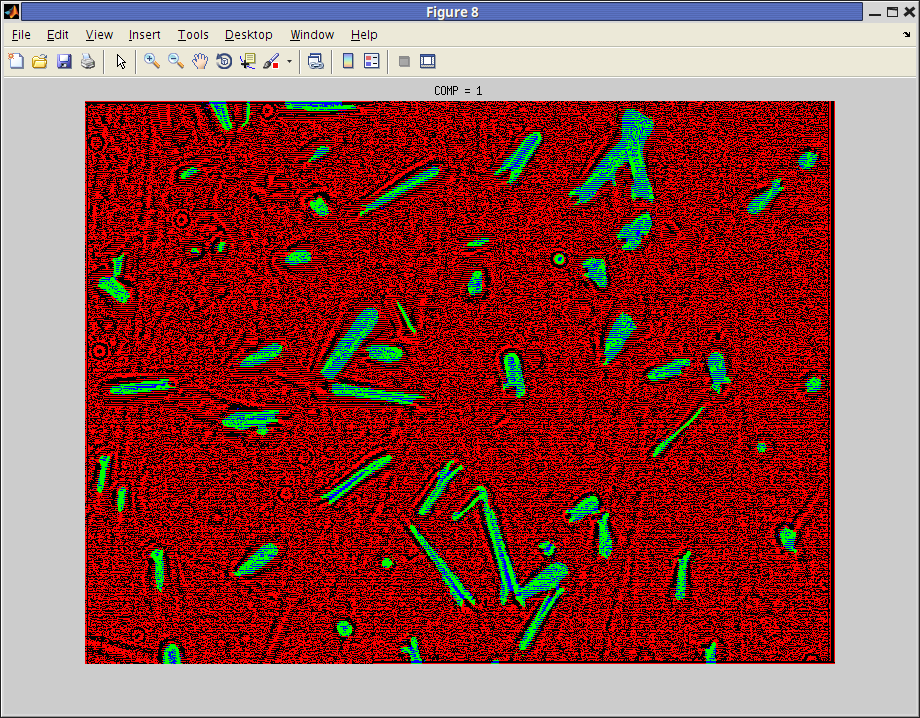}}}
    ~
    \subfloat[Window \texttt{COMP2}.]
    {\label{subfig:test2comp2}
    {\includegraphics[width=0.32\textwidth]{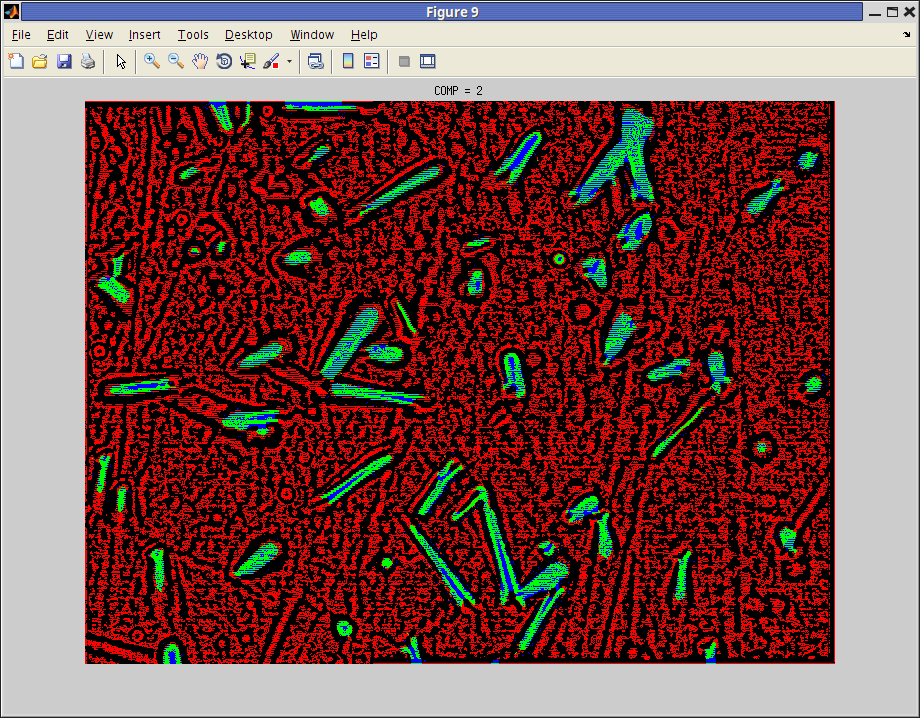}}}
    ~
    \subfloat[Window \texttt{COMP3}.]
    {\label{subfig:test2comp3}\
    {\includegraphics[width=0.32\textwidth]{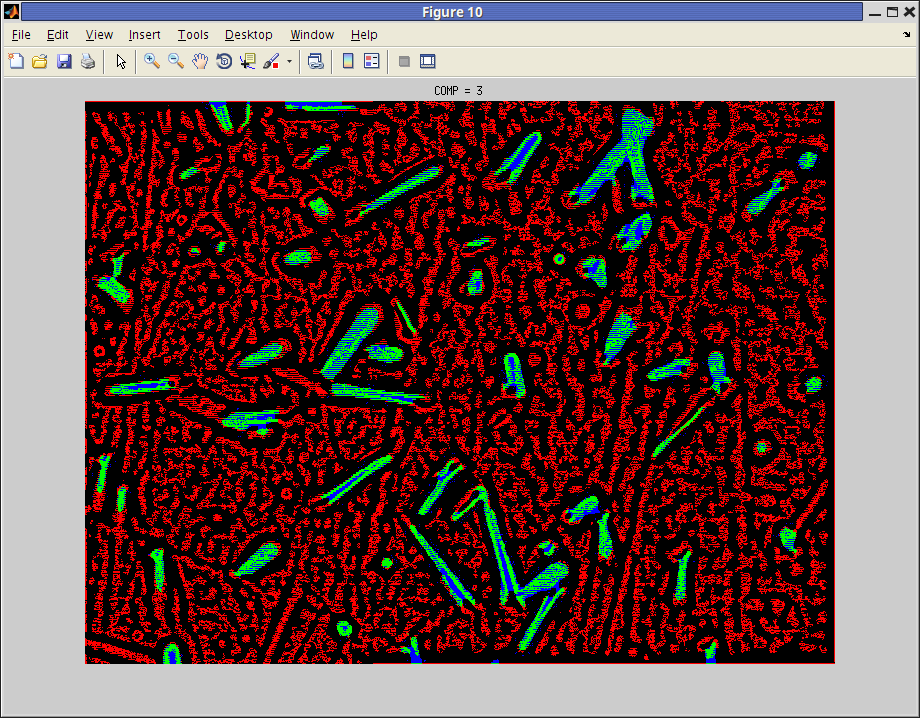}}}\\
    \caption{Windows presenting Jansen-MIDAS's application results on the
    GUI version, using \texttt{test2.jpg} and \texttt{test2GT.jpg}. The
    MLSOS application returns three image sets: details (\texttt{D}),
    segmentation (\texttt{R}) and comparison (\texttt{COMP}). The
    segmentation levels are $1$ (first) and $3$ (last), returning three
    images in each set.}
    \label{fig:test2results}
\end{figure*}

When the user inputs the GT image, the software presents it and the
segmentation starts. Then Jansen-MIDAS shows a window containing MCC
values for each segmentation (Figure \ref{fig:test2mcc}) and the windows
presenting \texttt{D}, \texttt{R} and \texttt{COMP}, according to the
user's choices (Figure \ref{fig:test2results}).

In this example, the check boxes \textit{Show D} and \textit{Show R} were
checked. Therefore, Jansen-MIDAS returns the image sets of starlet detail
levels (\texttt{D}) and MLSS results (\texttt{R}). When MLSOS is applied,
the images referring to the comparison between MLSS results and the GT
corresponding to the input photomicrograph (\texttt{COMP}) is presented
automatically. These results can be stored using the \textit{Save} menu
or button on each window. When the software ends its processing, the
welcome text changes its information to \textit{Done}.

\section{Discussion}
\label{sec:discussion}

Most of the segmentation methods demand human intervention at
some point of their execution. For example, the user of \cite{PINTO2013}
needs to select between 60 and 70 \% of well-binarized cells, and then
the method continues the processing. Also, \cite{GREENBLUM2014} uses
texture-based features for providing a coarse segmentation of dendritic
structures from \textit{C. elegans}, which is improved after by
post-processing.

Automatic segmentation methods benefit their users, which employ less
effort on image processing. Besides, multi-level segmentation methods
based on isotropic undecimated wavelets may enhance the segmentation
process by offering a set of detail coefficients for each wavelet
decomposition level. Based on these ideas, the algorithms implemented in
Jansen-MIDAS were used previously on the study of two different materials:

\textbf{Gold nanoparticles reduced on natural rubber membranes.}
The original MLSS algorithm implemented in Jansen-MIDAS was used for
separating gold nanoparticles in scanning electron microscopy (SEM)
photomicrographs reduced on natural rubber membranes \cite{DESIQUEIRA2014a},
thus being able to estimate the amount of synthesized gold nanoparticles
contained on the surface of these samples \cite{DESIQUEIRA2014c}. The
amount of nanoparticles within a sample can be estimated combining Mie's
theory and ultraviolet-visible spectroscopy \cite{HAISS2007}, a laborious
approach. Moreover, it is difficult to define the stoichiometry and
necessary parameters to estimate the density, i.e. the
concentration/distribution of synthesized nanoparticles over a substrate,
of organic substances developed using green chemistry.

\textbf{Fission tracks on the surface of epidote crystals.}
The derivative MLSS algorithm implemented in Jansen-MIDAS was used for
separating fission tracks in photomicrographs obtained from the
surface of epidote crystals \cite{DESIQUEIRA2014b}, which contain small
ROI. Usually these tracks are counted manually on an optical microscope.
There are commercial systems which perform this operation; for example,
\cite{GLEADOW2009} describes an automatic method for counting fission
tracks, based in two photomicrographs obtained from transmitted and
reflected lights. These images are binarized and their intersection
generates a coincidence mapping, which is used for the track analysis.
A commercial system based on this method is available; however, the
results acquired using this system often needs to be manually adjusted
by the operator, being more time consuming than the usual measure
\cite{ENKELMANN2012}. Jansen-MIDAS's application on the surface of
epidote crystals had an accuracy higher than 89 \%, and our approach can
be extended to be an open alternative to these systems.

\section{Conclusion}
\label{sec:conclusion}

In this article we presented Jansen-MIDAS, a software developed to provide
Multi-Level Starlet Segmentation (MLSS) and Multi-Level Starlet Optimal
Segmentation (MLSOS) techniques. These methods are based on the starlet
transform, an isotropic undecimated wavelet, in order to determine the
location of objects in photomicrographs.

MLSS uses the addition of detail levels obtained from applying the
starlet transform. There are two possible algorithms for MLSS: the input
image may be subtracted (original algorithm) or not (derivative
algorithm) from the sum of the detail coefficients. MLSOS, in its turn,
chooses the optimal segmentation level from MLSS based on the Matthews
correlation coefficient (MCC), which establishes the comparison between
the set of training images and their ground truths.

Jansen-MIDAS is an open-source software released under the GNU General
Public License. Its previous versions were used in the study of two
different materials, returning an accuracy higher than 89 \% in both
applications. Jansen-MIDAS is presented in two versions: a text-based
version, available for GNU Octave, and a graphical user interface (GUI)
version, compatible with MathWorks MATLAB, which can be employed on the
segmentation of several types of images, becoming a reliable alternative
to the scientist.

\section*{Acknowledgements}

The authors would like to acknowledge the São Paulo Research Foundation
(FAPESP), grants \# 2007/04952-5, 2009/04962-6, 2010/03282-9,
2010/20496-2, and 2011/09438-3.

\clearpage

\section*{References}

\bibliographystyle{ieeetr}
\bibliography{references}

\end{document}